\documentclass[10pt,twocolumn,letterpaper]{article}

\usepackage{iccv}              % To produce the CAMERA-READY version
\usepackage[accsupp]{axessibility} %
\definecolor{iccvblue}{rgb}{0.21,0.49,0.74}
\usepackage[pagebackref,breaklinks,colorlinks,allcolors=iccvblue]{hyperref}
\usepackage{multirow}
\usepackage{pifont}
\def\paperID{7721} % *** Enter the Paper ID here
\def\confName{ICCV}
\def\confYear{2025}

\title{Learning Visual Proxy for Compositional Zero-Shot Learning}

\author{Shiyu Zhang$^{1}$, Cheng Yan$^{1}$\thanks{Corresponding author.}, Yang Liu$^{2}$, Chenchen Jing$^{3}$, Lei Zhou$^{4}$, Wenjun Wang$^{1}$\\
{$^1$Tianjin University}\ \ {$^2$Zhejiang University}\\ {$^3$Zhejiang University of Technology}\\
{$^4$Hainan University}\\
{\tt\small \{zsy\_202366, yancheng\_work, wjwang\}@tju.edu.cn}\\[-3pt]
{\tt\small yangliu9610@zju.edu.cn}\\[-3pt]
{\tt\small jingchenchen@zjut.edu.cn}, {\tt\small leizhou@hainanu.edu.cn}
}
\begin{document}
\maketitle
\begin{abstract}
Compositional Zero-Shot Learning (CZSL) aims to recognize novel attribute-object compositions by leveraging knowledge from seen compositions. Current methods align textual prototypes with visual features via Vision-Language Models (VLMs), but suffer from two limitations: (1) modality gaps hinder the discrimination of semantically similar pairs, and (2) single-modal textual prototypes lack fine-grained visual cues. In this paper, we introduce \textbf{Visual Proxy} Learning, a method that reduces modality gaps and enhances compositional generalization. We initialize visual proxies for attributes, objects, and their compositions using text representations and optimize the visual space to capture fine-grained cues, improving visual representations. Additionally, we propose \textbf{Cross-Modal Joint Learning (CMJL)},  which imposes cross-modal constraints between the text-image and fine-grained visual spaces, improving generalization for unseen compositions and discriminating similar pairs. Experiments show state-of-the-art performance in closed-world scenarios and competitive results in open-world settings across four CZSL benchmarks, demonstrating the effectiveness of our approach in compositional generalization. The code will be available at \url{https://github.com/codefish12-09/VP_CMJL}.
\end{abstract}    
\section{Introduction}
\label{sec:intro}
In human cognition, the ability to recombine existing concepts to form new ones is essential for quickly acquiring new knowledge, a skill known as compositional generalization. Similarly, in computer vision, this ability is crucial for advancing models' adaptability to novel situations, which has led to the development of Compositional Zero-Shot Learning (CZSL) \cite{czsl17,czsl19}. The goal of CZSL is to enable models to decompose and recombine concepts learned from seen compositions of attributes and objects, and then generalize to unseen compositions, thereby addressing the zero-shot image classification task. For instance, if a model is trained on compositions such as green clothes and red apples, it should be capable of recognizing novel compositions, like red clothes and green apples, during testing.

\begin{figure}[!t]
  \centering
   \includegraphics[width=0.9\linewidth]{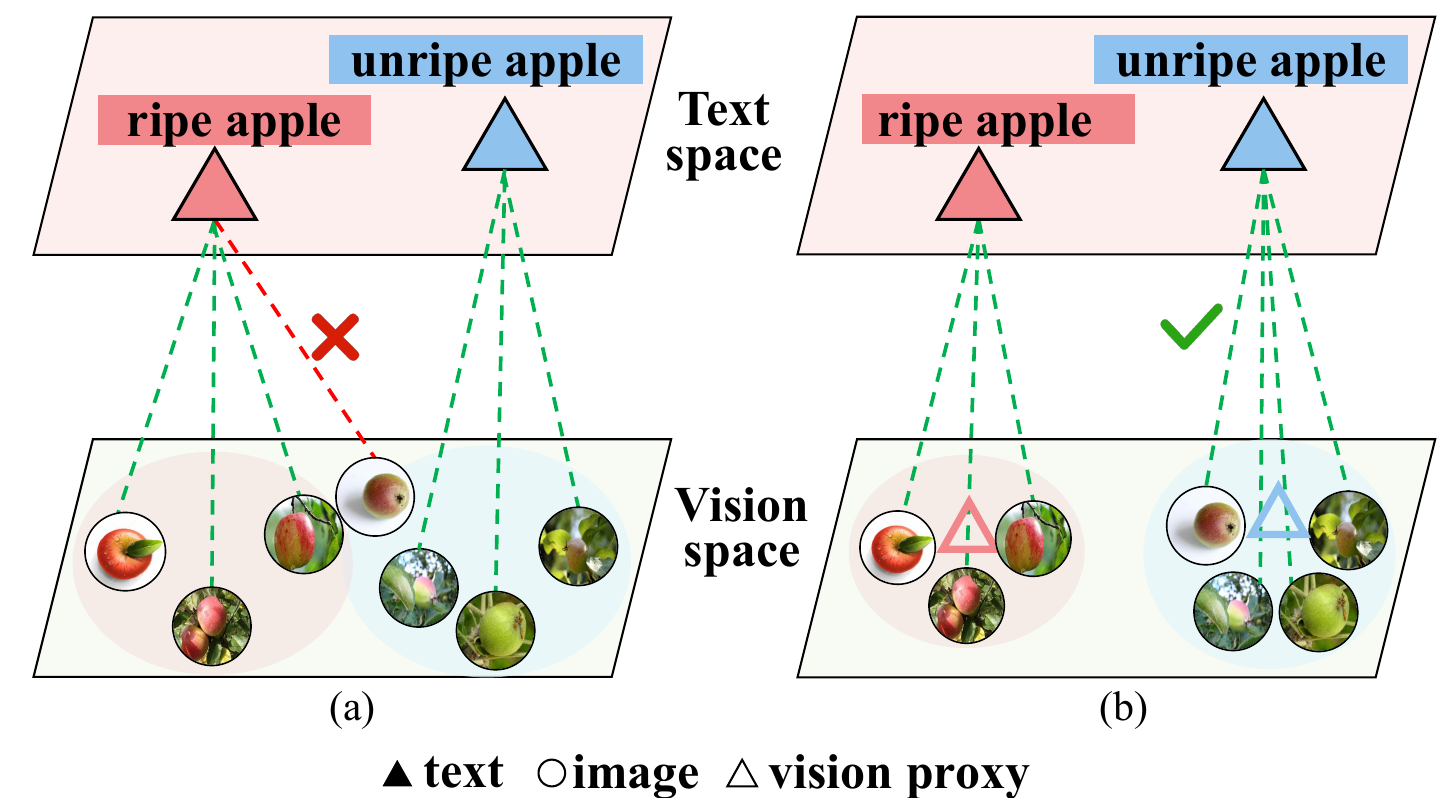}
   % \caption{(a) Given the two categories of samples, "ripe apple" and "unripe apple," the inter-class boundaries in the visual space become blurred due to the modality gap and the coarse granularity of textual prototypes. This results in classifiers aligned with textual prototypes being prone to misclassifying similar samples (slightly red unripe apples) into the wrong category. (b) The introduction of the visual proxy not only aligns the text and image modalities more effectively but also incorporates fine-grained image features, making the visual space more compact and separable, thereby ensuring correct classification of similar samples.}
   \caption{(a) The modality gap and coarse granularity of textual prototypes blur inter-class boundaries, leading to misclassification of similar samples (e.g., slightly red unripe apples). (b) The visual proxy improves modality alignment, integrates fine-grained image features, and makes the visual space more compact and separable, ensuring correct classification of similar samples.}
   % \caption{(a) Given an image of an ancient clock, which primarily carries the attribute "ancient" but also includes the semantic detail "small," the modality gap in VLM, diversity in image space, and fine-grained differences negatively affect the classification result. (b) To address this, we introduce prototypes in both modalities to improve modality alignment, facilitate fine-grained feature learning, and enable accurate classification during inference.}
   \label{fig:intro}
   % A comparison example between text prototype-based classification and visual prototype-based classification. Given a test image of an ancient clock containing the main attribute "ancient" but also including the semantic information "small." Due to the gap between the vision and textual modalities, the test image is divided into small blocks. However, the introduction of visual prototypes enables accurate classification by emphasizing the distance from the image to the visual prototype of its class over other classes.
   %%%%\vspace{-0.5cm}
\end{figure}

Thanks to the advancements in large pre-trained vision-language models like CLIP \cite{clip}, recent research has made significant progress by leveraging CLIP and exploring various approaches to enhance its capabilities. These approaches include designing task-specific prompts \cite{csp23, PCVL, glpcol24, lu2023drpt, PLID}, decomposing compositional text features \cite{DFSP}, and developing modal fusion methods \cite{trokia24} to narrow the modality gap. Such efforts have improved the alignment between compositional labels and image features, enabling CLIP to perform effectively in CZSL tasks. However, several challenges and inherent problems in the CZSL task still deserve attention. 

\textbf{Persistent Modality Gap.} Despite efforts to narrow the modality gap between vision and language in existing methods, it persists as a fundamental challenge in CZSL. While prior work \cite{liang2022mind} demonstrates that the gap can be partially reduced, complete elimination remains unattainable, particularly under the current task setup where classification depends critically on \textit{top-1 cross-modal retrieval}. In such scenarios, the distance between true cross-modal pairs may exceed that of false pairs, leading to misclassifications directly attributable to the modality gap (see \cref{fig:intro}(a)). This problem is further amplified by the fine-grained nature of compositional semantics, where subtle visual distinctions—such as texture variations between 'ripe' and 'unripe' apples—are semantically critical yet notoriously difficult to align with their textual counterparts.  

\textbf{Limited Discriminability of Textual Prototypes.} Existing VLM-based CZSL methods learn textual prototypes to address label scarcity, yet these prototypes inherently lack the fine-grained visual details crucial for distinguishing similar compositions. While each visual category contains diverse image instances (e.g., apples in varying shapes or lighting conditions), textual prototypes are derived from a single compositional label, resulting in a semantic-visual asymmetry. This mismatch prevents prototypes from capturing subtle distinctions (e.g., 'ripe' vs. 'unripe' apples), leading to ambiguous decision boundaries in the visual space (Fig. \ref{fig:intro}(a)). As shown in prior studies \cite{qian2024intra}, relying solely on textual prototypes fails to approach the optimal performance for visual tasks.

% 现有所有基于VLM的CZSL的方法，考虑到零样本分类中标签缺失问题，均致力于得到一个丰富的文本中心以代替类别标签。虽然文本代理在CZSL任务中很好获得，但其缺乏细粒度特征，而在CZSL中，这种细粒度特征对区分相似组合对是极其重要的，These features reside in the visual space, as each category contains multiple image instances, whereas only a single compositional text label is available. 有相关研究证明，仅仅依靠文本代理无法获得视觉任务的最优解。
% 因此，在视觉空间中学习类别中心是有必要的。但考虑到视觉模态不同于文本模态，视觉特征通常具有高度的多样性和变异性，同一类别的对象在不同场景、视角、光照条件下可能表现出截然不同的视觉特征，其中心较难获得，而文本模态已较为成熟。为此，我们首次在CZSL中提出视觉代理的概念，即利用文本模态引导视觉中心的学习。

% (2) Existing methods primarily focus on inter-modal feature alignment and the use of various prompts \cite{csp23, PCVL, lu2023drpt} to enhance text features, yet they often overlook the importance of the visual modality. In CZSL tasks, where compositional text labels consist of multiple attributes and objects, the model must capture both the diversity and fine-grained details of categories. These features reside in the visual space, as each category contains multiple image instances, whereas only a single compositional text label is available. Therefore, relying solely on textual prototypes is insufficient to fully represent these visual features. For instance, as shown in \cref{fig:intro}(b), the image of an 'ancient clock' is very similar to that of a 'small clock,' and distinguishing between them is difficult without fine-grained visual information.

Therefore, learning a class center in the visual modality is essential for CZSL. However, directly learning such centers is challenging due to the high variability of visual features (e.g., viewpoint and lighting variations), while the textual modality offers a more compact and semantically structured space. To overcome this challenge, we introduce the novel concept of \textbf{Visual Proxy} in CZSL, which leverages the well-established textual modality to guide the learning of visual centers. Building on this idea, we propose a \textbf{Cross-Modal Joint Learning (CMJL)} strategy that enables the collaborative optimization of textual prototypes and visual proxies, capitalizing on their complementary strengths while jointly facilitating composition pair recognition during the testing phase. This approach not only preserves the strong generalization capabilities of CLIP but also enhances discriminability in the visual space. In summary, our Visual Proxy-Based Cross-Modal Joint Learning(\textbf{VP-CMJL}) achieves two key objectives:
(1) \textbf{Enhancing modality alignment:} By introducing visual proxies, semantic-level alignment from textual prototypes to the visual space can be achieved, alleviating semantic projection bias.
(2) \textbf{Preserving Visual Granularity:} Visual proxies preserve fine-grained details through contrastive learning, enabling precise discrimination of visually similar categories, and making the visual feature space more compact and separable, as shown in \cref{fig:intro}(b).

Our main contributions are summarized as follows: 
\begin{itemize}
    \item \textbf{Visual Proxy.}  
    We introduce a novel concept of visual proxy, the first to be applied in the CZSL task, as a learnable class center in the visual modality guided by the text modality. This concept not only enhances modality alignment but also captures fine-grained image features, improving the discrimination of similar compositions.
    \item \textbf{Cross-Modal Joint Learning.}  
    We propose an effective Cross-Modal Joint Learning strategy that simultaneously optimizes textual prototypes and visual proxies through cross-modal constraints, leveraging the complementary strengths of both modalities to jointly predict compositions, while enhancing the model's generalization and discriminative abilities.
    \item \textbf{Superior Performance on CZSL Benchmarks.} 
    Extensive experiments on MIT-States, UT-Zappos, C-GQA and VAW-CZSL demonstrate state-of-the-art performance, validating the effectiveness of visual proxies and cross-modal joint learning.
\end{itemize}

\section{Related Work}
\label{sec:relatedwork}
\textbf{Compositional Zero-Shot Learning.} Previous methods for CZSL can be categorized into two strategies: (1) Uniform feature representation learning, which establishes relationships between compositions and primitives (attributes and objects), embedding both images and compositions into a shared space for unseen composition prediction using a single classifier \cite{purushwalkam2019task,cgqa,czsl17,anwaar2022leveraging,li2022siamese}. 
% naeem2021learning
(2) Multi-branch feature learning, which employs parallel discriminative modules for attributes, objects, and compositions, emphasizing visual feature discriminability \cite{nagarajan2018attributes,karthik2022kg,kim2023hierarchical,wang2023learning,saini2022disentangling,li2020symmetry}. Recently, with the advancement of pre-trained Vision-Language Models (VLMs) and their strong generalization abilities, VLMs have been increasingly applied to CZSL \cite{csp23,PCVL,lu2023drpt,DFSP}. For instance, Nayak et al. \cite{csp23} used soft prompts to treat attributes and objects as learnable tokens, while Xu et al. \cite{glpcol24} constructed a graph \cite{wo2024graph} of object-attribute compositions. Lu et al. \cite{DFSP} decomposed states and objects in language features and integrated them with image features. Huang et al. \cite{trokia24} created three recognition branches to model attributes, objects, and compositions, aligning branch-specific prompt representations with visual features. Li et al. \cite{li2024context} proposed a context-based, diversity-driven specificity learning framework, considering attribute specificity levels. 

\noindent
\textbf{Visual Center Learning.} Previous work aims to obtain visual centers through prototypes by averaging image features within a single modality \cite{li2022siamese, ruis2021independent}. However, such approaches face inherent limitations: (1) Prototypes derived solely from visual features are sensitive to viewpoint and lighting variations, leading to poor generalization; (2) Uni-modal prototypes fail to capture fine-grained semantics of attribute-object compositions. To address these issues, we propose the novel concept of Visual Proxy, which introduces textual modality as semantic guidance for the first time. Through cross-modal joint learning, Visual Proxy dynamically calibrates visual centers, achieving a balance between generalizability and discriminability.
% Recently, some studies have introduced prototypes into CZSL\cite{li2022siamese,ruis2021independent}, where they obtain the prototypes by averaging image features, with the prototypes derived from a single modality. Our work differs from the aforementioned prototype-updating methods. We introduce textual and visual prototypes during the training phase and make them both learnable. By aligning the dual-modal prototypes with image features, we learn textual prototypes that represent broad concepts and visual prototypes that capture fine-grained features.
\section{Method}
% 正如section \cite{section1}所分析，CZSL任务要求模型不仅在未见组合上具备很好的泛化能力，同时也要求模型具备更细粒度的判别能力，以应对相似组合对判别易错的问题，为了应对这一挑战，我们提出了一种新的学习策略，名为双模态原型学习Dual-Modal Prototype Learning for CZSL (DMPL-CZSL)，它构建了文本原型和视觉原型联合学习的新范式。基于现有的属性、对象和组合对的三路预测方式[]，首先，在文本模态中，DMPL-CZSL构建关于属性、类别和组合对的可学习提示词，得到各自特征，进而获得文本原型。同时，DMPL-CZSL在视觉模态上，构建属性、类别和组合对的可学习视觉原型。然后，为了学习每个原语及组合对的最优文本和视觉原型，来提高在可见对和不可见对上利用双原型分类的有效性，我们设计了针对双原型的联合学习训练方法和适应于不同原型的图像特征解耦策略。我们提出的方法框架如图2所示。
\begin{figure*}[t]
  \centering
  \includegraphics[width=0.87\textwidth]{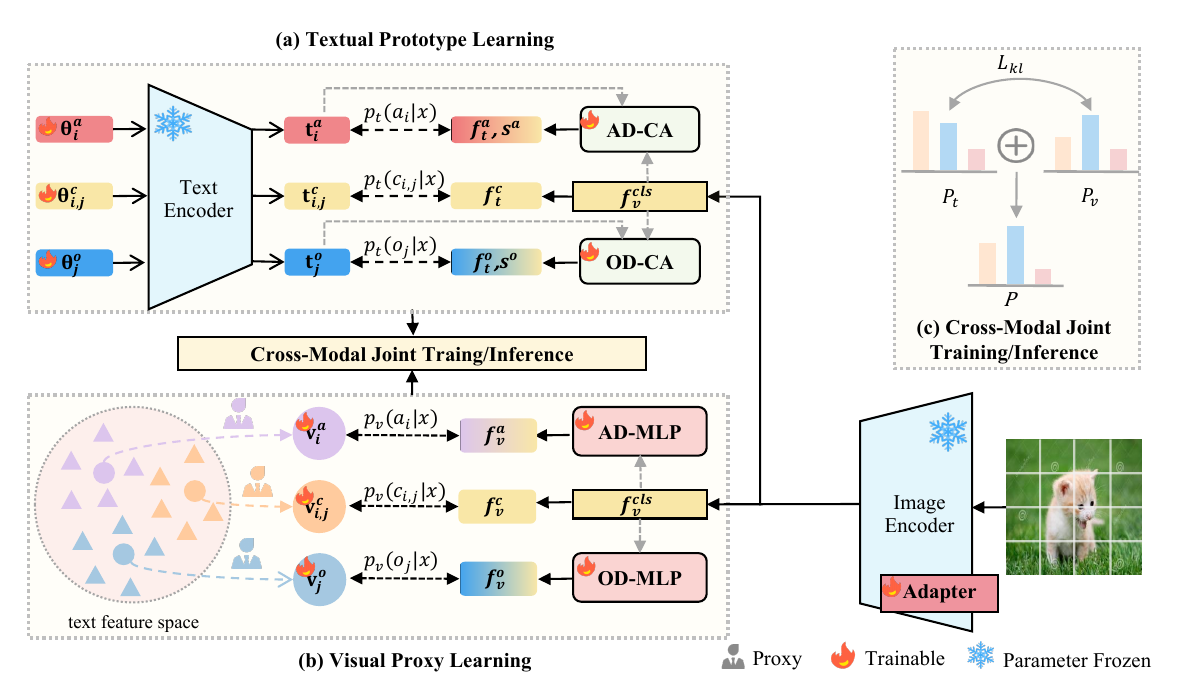}
  % \caption{The proposed CMPL framework primarily consists of textual and visual prototype learning components, each featuring three distinct pipelines: attribute, object, and composition, all aimed at feature and prototype learning. For intra-modal learning, a dedicated decomposition module is designed to improve the feature quality. For inter-modal learning, a dual-prototype joint training strategy is introduced, enabling the optimization of both textual and visual prototypes. During inference, the final probability score is obtained by summing the probabilities of the two prototypes.}
  \caption{The overview of VP-CMJL. (a) The Textual Prototype Learning Module utilizes a three-path framework and cross-modal image feature decoupling module to enhance semantic alignment between text and image. (b) The Visual Proxy Learning Module guides visual proxy learning using the text modality, aggregates hierarchical visual features, and preserves fine-grained discriminative information. (c) The Cross-Modal Joint Learning module employs a KL divergence constraint to collaboratively optimize textual prototypes and visual proxies. During the inference phase, both are combined to achieve composition pair prediction.}
  % \caption{\textbf{Overview of the proposed CMPL.}}
  % The overall framework consists of three paths for attribute, object, and composition prediction, divided into three parts: text prototype learning, visual prototype learning, and joint training and testing. Given an image, global features are extracted and fed into the text prototype learning module and the visual prototype learning module separately. For text prototype learning: the image features are decoupled using a cross-attention decoupling module, aligned with the textual representation of learnable prompting words to learn the text prototypes of the three paths. For visual prototype learning: MLP is used to transform the image features, aligned with learnable visual prototypes to learn visual prototypes with fine-grained features. The KL divergence is calculated on the category probability distributions obtained from the two modules to achieve dual-prototype joint learning. During testing, the sum of the probabilities of the two prototypes is used as the final probability score.
  \label{model}
  %\vspace{-0.5cm}
\end{figure*}
\subsection{Problem Formulation}
% 在组合零样本学习任务中，文本标签由属性和对象两部分组成。属性记为$a\in A$ ，类别记为$o \in O$，组合记为$y \in Y$，其中$Y=A \times O$。在整个组合标签空间中，划分为可见组合对集合$Y_S$和不可见组合对集合$Y_U，其中$Y_S \cap  Y_U=\Phi$。在训练阶段，给定训练集合$T={(x,y)|x \in X,y \in Y_s}$,其中，X代表图像空间。在测试阶段，模型需要从可见和不可见的组合对中对图像进行分类。在未见组合对的设定上，组合零样本学习分为closed-world和open-world两种设定[]。其中closed-world设定中，未见过的组合对$Y_U$作为先验知识给出，预测空间为$Y_pred=Y_S \cup Y_u$，而在open-world中，未见过的组合对为全部属性和对象的排列组合，预测空间为$Y_pred=A \times O$。

In CZSL, compositional labels consist of attributes and objects. Attributes are denoted as $a \in \mathcal{A}$, objects as $o \in \mathcal{O}$, and compositions as $c \in \mathcal{C}$, where $\mathcal{C} = \mathcal{A} \times \mathcal{O}$. The compositional label space is divided into seen compositions $\mathcal{C}_s$ and unseen compositions $\mathcal{C}_u$, with $\mathcal{C}_s \cap \mathcal{C}_u = \varnothing$. The training set is defined as $\mathcal{T} = \{{(x, c) |x \in \mathcal{X}, c \in \mathcal{C}_s \}}$, where $\mathcal{X}$ represents the image space. During testing, the model must classify images from both seen and unseen compositions. CZSL can be categorized into closed-world and open-world settings based on the treatment of unseen compositions \cite{mancini2021open}. In the closed-world setting, unseen compositions $\mathcal{C}_u$ are provided as prior knowledge, and the prediction space is $\mathcal{C}_{pred} = \mathcal{C}_s \cup \mathcal{C}_u$. In contrast, in the open-world setting, unseen compositions include all possible attribute-object compositions, and the prediction space is $\mathcal{C}_{pred} = \mathcal{A} \times \mathcal{O}$.

\subsection{Basic Framework}
\label{Basic Framework}
% 为保证在未见组合对上的泛化能力，我们跟随前人的工作，利用CLIP的编码器进行特征提取，并采用CZSL任务中通用有效的三路预测框架。
To ensure generalization ability on unseen compositions, we follow previous works by utilizing frozen CLIP\cite{clip} and adapters\cite{chen2022adaptformer} for feature extraction, and adopt the commonly effective three-path(attribute, object and composition) prediction framework \cite{trokia24,li2024context} in CZSL tasks.

\noindent
\textbf{Visual Representation.} We leverage the CLIP image encoder $E_v$ to extract the global image feature $f_v^{\text{cls}}$ from the input image $x$, which serves as the image feature $f_v^c$ for the composition branch. To achieve feature disentanglement, we employ two independent MLPs to process $f_v^c$, yielding the \textit{attribute-oriented} image feature $f_v^a$ and the \textit{object-oriented} image feature $f_v^o$.

\noindent
\textbf{Prompt Representations.} Following previous work, \cite{csp23,coop,trokia24} we first construct the attribute prompt $\theta_{i}^a = [p_0^a, \dots, p_m^a, w_i^a]$, the object prompt $\theta_{j}^o = [p_0^o, \dots, p_m^o, w_j^o]$, and the composition prompt $\theta_{i,j}^c = [p_0^c, \dots, p_m^c, w_i^a, w_j^o]$. Here, $p_{0:m}^a$, $p_{0:m}^o$, and $p_{0:m}^c$ are prefixes initialized with the phrase \textit{``a photo of''}, while $w_i^a$ and $w_j^o$ represent the vocabulary tokens for the attribute $a_i$ and the object $o_j$, respectively. All components of the prompts are set as learnable parameters. Subsequently, we utilize the CLIP text encoder $E_t$ to obtain the attribute text feature $t_i^a$, the object text feature $t_j^o$, and the composition text feature $t_{i, j}^c$.

%现有的CZSL方法均沿用CLIP中文本对图像这种点对点的分类范式，忽略模态不平衡问题和视觉模态所具备的细粒度特征，导致相似组合对分类错误。我们的方法首次提出基于三路的双模态原型联合学习框架，如图2所示。在训练阶段，我们设计不同的解耦模块分别学习文本原型和视觉原型，并提出联合训练策略促进双原型的学习，最终得到最优的文本原型、视觉原型和紧凑的图像特征。在测试阶段，我们结合图像特征与双原型的预测概率进行组合对识别。该方法既增强了相似组合对的判别能力，又提升了在未见组合对的泛化性。
\noindent
\textbf{Our Main Idea.} Existing CZSL methods follow the CLIP classification paradigm through text-image matching, but the modality gap causes confusion between similar composition pairs, and relying solely on textual prototypes as class labels lacks fine-grained features. To address these issues, we introduce text-guided visual proxies in the visual space and propose a Cross-Modal Joint Learning strategy, referred to as \textbf{VP-CMJL}, as shown in \cref{model}, which consists of the following two key innovative modules:
\begin{itemize}

\item \textbf{Text-Guided Visual Proxy Learning:}
We introduce \textbf{visual proxies}—learnable centers in the visual space explicitly guided by textual semantics. Unlike traditional visual centers, visual proxies are dynamically optimized through cross-modal alignment, with text features serving as anchors to regularize the learning process. To enable better guidance of visual proxy learning by the text modality, we also design a corresponding textual prototype learning module, further enhancing modality alignment.
 \item \textbf{Cross-Modal Joint Learning:}  
During training, \textbf{CMJL} jointly optimizes visual proxies and textual prototypes using KL divergence, enabling the collaborative learning of both modalities. During inference, we compute and combine two compatibility scores: (1) between image features and textual prototypes for semantic alignment, and (2) between image features and visual proxies for fine-grained discrimination. The final prediction is a weighted fusion of these scores, dynamically balancing cross-modal consistency and visual specificity to enhance robustness.
\end{itemize}

\subsection{Text-Guided Visual Proxy Learning}
\subsubsection{Textual Prototype Learning}
The textual prototypes capture class-specific semantics at three levels (attributes, objects, compositions), following the multi-branch design in \cite{trokia24} to enhance generalization. For visual features, we introduce attribute cross-modal decoupling modules (\textbf{AD-CA}) and object cross-modal decoupling modules (\textbf{OD-CA}), aligning them with the corresponding textual prototypes to improve cross-modal consistency. Furthermore, we propose a novel probability calculation method that integrates attention scores.
The text representations are treated as textual prototypes, denoted as follows:
% \begin{equation}
% t^a=[t_1^a,t_2^a,...,t_{|\mathcal{A}|}^a], \quad t^o=[t_1^o,t_2^o,...,t_{|\mathcal{O}|}^o], 
% \end{equation}
% % \begin{equation}
% % t^o=[t_1^o,t_2^o,...,t_{|\mathcal{O}|}^o], 
% % \end{equation}
% \begin{equation}
% t^c=[t_{1,1}^c,t_{1,2}^c,...,t_{1,j},...,t_{|\mathcal{C}_s|}^c],
% \end{equation}
\begin{equation}
t^y = [t_1^y, t_2^y, \dots, t_{|\mathcal{Y}|}^y], 
\end{equation}
where $y \in \{a,o,c\}, \mathcal{|Y|} \in \{\mathcal{|A|},\mathcal{|O|},\mathcal{|C}_s|\}$. $|\mathcal{A}|$ and $|\mathcal{O}|$ denote the number of attributes and objects, respectively, $|\mathcal{C}_s|$ indicates the number of seen compositions during the training phase.

\noindent
\textbf{Cross-Modal Decoupling Module.} Considering the gap between text and image, simple MLPs are insufficient for disentangling image features and aligning them with textual prototypes. To address this, we propose cross-modal decoupling modules, implemented as multi-head cross-attention \cite{chen2021crossvit}, for the attribute and object branches, denoted as \textbf{AD-CA} and \textbf{OD-CA}. We focus on the attribute branch here, with the object branch following a similar process. Given the composition image feature $f_t^c=f_v^{cls} \in \mathbb{R}^{d}$ and attribute text prototypes $t^a \in \mathbb{R}^{|\mathcal{A}| \times d}$ as inputs, the query, key, and value are formulated as:
% \begin{equation} 
% q = f^c W^q, k = t^a W^k, v = t^a W^v,
% \end{equation}
% where $W^q$, $W^k$, and $W^v$ $\in \mathbb{R}^{d \times d_k}$ are the parameter matrices, with $d_k = d/h$, where $h$ is the number of attention heads. 
% The attention score $s^a$ between the image feature $f^c$ and the attribute text prototype $t^a$ is given by:
% \begin{equation}
% \text{$s^a$} = \mathrm{softmax}\left(\frac{qk^\top}{\sqrt{d_k}}\right) \in \mathbb{R}^{1 \times n} \quad 
% \end{equation}
% The decoupled image feature $f^a$ is obtained as:
% \begin{equation}
% f^a = \text{$s^a$} \cdot v \in \mathbb{R}^{d_k}. \quad
% \end{equation}
% 公式1: 线性投影
\begin{equation} 
    Q_a = f_t^c W_a^q,\quad  K_a = t^a W_a^k, \quad V_a = t^a W_a^v,
\end{equation}
where $W_a^q$, $W_a^k$, and $W_a^v$ $\in \mathbb{R}^{d \times d_k}$ are the parameter matrices, with $d_k = d/h$, where $h$ is the number of attention heads. 

The attention score $s^a$ between the image feature $f_t^c$ and the attribute text prototype $t^a$ is given by:
\begin{equation} 
    s^a = \frac{1}{h} \sum_{i=1}^h \mathrm{softmax}\left( \frac{Q_{a,i} {K}_{a,i}^\top}{\sqrt{d/h}} \right).
\end{equation}

The multi-head output is obtained by concatenating and linearly projecting:
\begin{equation} 
    O_a = \mathrm{Concat}\left( s_1^aV_{a,1},\ \dots,\ s_h^aV_{a,h} \right) W_a^o,
\end{equation}
where $W_a^o \in \mathbb{R}^{d \times d}$ is the weight matrix.

The output $O_a$ is then passed through a feed-forward network, followed by layer normalization and a residual connection to obtain the decoupled image feature $f_t^a$:
\begin{equation} 
    f_t^a = f_t^c + \mathrm{FFN}_a\left( \mathrm{LN}\left( f_t^c + O_a \right) \right).
\end{equation}

After the \textbf{AD-CA} module, the attribute image feature $f_t^a$ and attention score $s^a$ are derived from the image feature $f_t^c$. Similarly, the object image feature \( f_t^o \) and attention score $s^o$ can be obtained from \textbf{OD-CA}.
% Considering the mutual dependence between the attribute and object branches, which cannot be fully decoupled, we retain the attribute branch attention score $s^a$ and use it in subsequent probability prediction calculations. 

\noindent
\textbf{Attention Score-Based Probability Calculation.}
% 在获得各分支的文本原型和图像特征后，我们计算属性、对象和组合的概率。与以往的研究不同\cite{trokia24,li2024context}，考虑到在czsl中，属性和对象是相互纠缠的，我们在使用分解的图像特征和相应的文本原型单独计算属性和对象概率分数的同时，也结合了原始图像特征与属性、对象文本原型的注意力权重，以加强对属性和对象的预测。具体概率计算如下：
After obtaining the textual prototypes and image features for each branch, we calculate the probabilities for attributes, objects, and compositions. Unlike previous studies \cite{trokia24, li2024context}, which treat the attribute and object branches separately, we account for the entanglement between attributes and objects in CZSL. For both the attribute and object branches, we incorporate the attention weights \(s^a\) and \(s^o\) between the original image features and textual prototypes into the traditional text-image similarity, enhancing the prediction of attributes and objects. The specific probability calculation is as follows:

The probability values of the Attribute and Object branches:
\begin{equation}
p_t({y_i}|x) = \frac{\exp((f_t^y \cdot t_i^y + s_i^y) / \tau_t)}{\sum_{k=1}^{|\mathcal{Y}|} \exp((f_t^y \cdot t_k^y + s_k^y) / \tau_t)},
\end{equation}
where $y \in \{{a,o}\}, |\mathcal{Y}| \in \{{|\mathcal{A}|,\mathcal|{O}|\}}$, \( y_i \) represents the specific element in the set for each case (i.e., \( a_i \) and \( o_j \)).
% \begin{equation}
% p_t({a_i}|x)=\frac{\exp((f_t^a\cdot{t}_i^a+{s_i}^a)/\tau)}{\sum_{k=1}^{|\mathcal{A}|}\exp((f_t^a\cdot {t}_k^a+{s_k}^a)/\tau)},
% \end{equation}
% \begin{equation}
% p_t({o_i}|x)=\frac{\exp(({f_t}^o\cdot{t}_i^o+{s_i}^o)/\tau)}{\sum_{k=1}^{|\mathcal{O}|}\exp(({f_t}^o\cdot{t}_k^o+{s_k}^o)/\tau)},
% \end{equation}

The probability values of the Composition branch:
\begin{equation}
p_t(c_{i,j}|x)=\frac{\exp(f_t^c\cdot{t}_{i,j}^c/\tau_t)}{\sum_{k=1}^{|\mathcal{C}_s|}\exp({f_t}^c\cdot{t}_k^c/\tau_t)},
\end{equation}
where $\tau_t \in \mathbb{R}$ denotes the temperature parameter, which is pre-trained in CLIP. Subsequently, we compute the cross-entropy loss for each branch as follows:
\begin{equation}
\mathcal{L}_t^y = -\frac{1}{|\mathcal{X}|} \sum_{x \in \mathcal{X}} \log p_t(y | x),
\end{equation}
where $y \in \{{a,o,c}\}$.
% \begin{equation}
%     \mathcal{L}_t^a =-\frac1{|\mathcal{X}|}\sum_{x\in\mathcal{X}}\log p_t(a|x), 
% \end{equation}
% \begin{equation}
%     \mathcal{L}_t^o =-\frac1{|\mathcal{X}|}\sum_{x\in\mathcal{X}}\log p_t(o|x),
% \end{equation}
% \begin{equation}
%     \mathcal{L}_t^c =-\frac1{|\mathcal{X}|}\sum_{x\in\mathcal{X}}\log p_t(c|x),
% \end{equation}
Therefore, the total loss for the textual prototypes learning module $\mathcal{L}_t$ is defined as:
\begin{equation}
\mathcal{L}_t=\gamma_{ao}(\mathcal{L}_t^a+\mathcal{L}_t^o)+\gamma_{c}\mathcal{L}_t^c.
\end{equation}
The parameters \(\gamma_{ao}\) and \(\gamma_{c}\) are hyperparameters, and the analysis can be found in the supplementary materials.
\subsubsection{Visual Proxy Learning}
\begin{figure}[htbp]
  %\vspace{-0.35cm}
  \centering
   \includegraphics[width=0.8\linewidth]{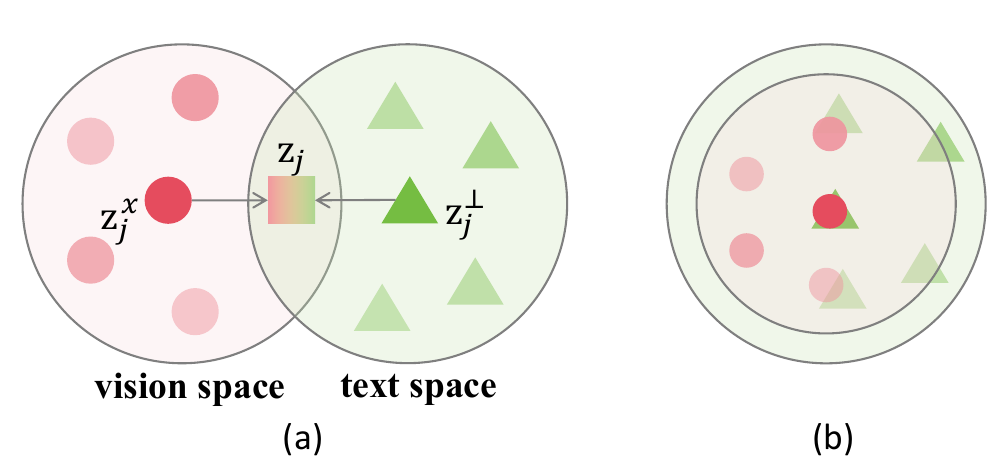}
   %\vspace{-0.2cm}
   \caption{(a) Illustration of the visual and text spaces, where the class prototype \( z_j \) resides in the overlapping region. (b) Ideal scenario where the visual space is fully covered by the text space.}
   \label{set}
   %\vspace{-0.3cm}
\end{figure}

% 上述文本原型学习模块，旨在通过缩短图像模态和文本模态之间的gap，来学习到一个类原型。如图3(a)所示，该类原型处在图像空间和文本空间重叠的区域，其可被分解为:
% 其中$\mathbf{z}_j^x$处于视觉空间，捕获了视觉模态的信息。$\mathbf{z}_j^\perp$位于与视觉空间正交的子空间中，包含了文本模态特有的信息，是在视觉模态无法直接表达的内容。由此可见，该类原型包含了来自视觉和文本的共同信息，但视觉模态和文本模态的差距依然存在，因此，放在交集区域的类原型也在一定程度上受限于这种差距的影响。考虑到CZSL任务的本质属于图像分类任务，该全部类原型的最优解一定分布在视觉空间中。只有当视觉空间全部被文本空间覆盖时，才能得到全部最优解，如图3(b)所示。由于两模态之间存在gap，不可能全覆盖，为此，我们在视觉模态中引入视觉原型，学习独属于视觉模态的细粒度类原型，以弥补文本原型学习不到的类原型。
\textbf{Why is the visual proxy needed?} The textual prototype learning module described above is designed to narrow the gap between the visual and textual modalities, facilitating the learning of a unified class center. Previous studies \cite{qian2024intra} have demonstrated that the class center exists in the overlapping region between the visual and text spaces. As shown in \cref{set} (a), $\mathbf{z}_j$ is the class center of CLIP, which can be presented by two features from both modalities:
\begin{equation}
\mathbf{z}_j=\sqrt{a}\mathbf{z}_j^x+\sqrt{1-a}\mathbf{z}_j^\perp, 
\end{equation}
where $\mathbf{z}_j^x$ is derived from the vision space and $\mathbf{z}_j^\perp$ shows the component from the orthogonal subspace such that $\mathbf{z}_j^{x\top}\mathbf{z}_j^\perp=0$, containing information unique to the textual modality. 
The class center $\mathbf{z}_j$ in the intersection area embodies shared information from both modalities. However, due to the gap between these modalities, the class center in the overlapping region is still influenced by this discrepancy, which means the ideal state of \cref{set}(b) is almost impossible to reach. Given that the core of the CZSL task is image classification, the optimal solution for all class centers must lie within the vision space. Thus the balance of learning should be appropriately tilted toward visual modality. Considering that the visual center is difficult to obtain, we propose leveraging the text modality to guide its learning. To this end, we introduce visual proxy learning.

% which contains fine-grained class prototypes specific to the visual domain, compensating for those that cannot be captured by the text prototype alone.

% Full optimal coverage through the text prototype can only be achieved when the visual space is entirely covered by the text space.

%However, since the modality gap can only be reduced, not fully eliminated, which means the ideal state of (b) is almost impossible to reach. Given that the core of the CZSL task is image classification, the optimal solution for all class prototypes must lie within the visual space. Full optimal coverage through the text prototype can only be achieved when the visual space is entirely covered by the text space. We introduce a visual prototype in the visual modality to learn fine-grained class prototypes specific to the visual domain, compensating for those that cannot be captured by the text prototype alone.
% Since the textual prototypes is not sufficient to capture fine-grained class features, we introduce visual prototypes in the visual modality. 

\noindent
\textbf{Visual Proxy Construction.} To ensure the model's generalization ability on unseen compositions, we introduce three-path visual proxy learning. We denote the attribute visual proxies as $v^a$ and the object visual proxies as $v^o$, as follows:
\begin{equation}
v^a=[v_1^a,v_2^a,...,v_{|\mathcal{A}|}^a], \quad v_i^a \in \mathbb{R}^{1\times d}
\end{equation}
\begin{equation}
v^o=[v_1^o,v_2^o,...,v_{|\mathcal{O}|}^o], \quad v_j^o \in \mathbb{R}^{1\times d}
\end{equation}
where $d$ represents the dimension of image features. 

% 考虑到CZSL任务中，同一属性在不同对象上所表现出的图像特征差异性较大，采用同一类别样本的平均图像特征来定义视觉原型\cite{zhang2024dual}存在局限性。
Given the inherent challenges of learning discriminative visual proxies via random initialization, which often leads to suboptimal local optima due to sparse training signals, we leverage the compact and semantically stable representations from pre-trained language models (PLMs). Specifically, we initialize visual proxies with text-derived embeddings:
\begin{equation}
    v_i^a=E_l(w_i^a),\quad v_j^o=E_l(w_j^o), 
\end{equation}
% 考虑到CZSL需要在测试阶段对未见过的组合进行分类，我们将属性视觉原型 $v^a$ 和对象视觉原型 $v^b$ 设置为可学习的，进而来获得组合视觉原型 \(v^c\) ：
where $E_l$ denotes the text encoder of PLMs like BERT\cite{kenton2019bert}, GPT\cite{radford2019language} or CLIP\cite{clip}. Importantly, our Appendix provides a comprehensive analysis of PLM selection for proxy initialization. 
The experiment shows that using CLIP's text feature as initialization yields the best results.

Given that CZSL requires the classification of unseen compositions during testing, we set the attribute visual proxies, $v^a$, and the object visual proxies, $v^o$, as learnable parameters, and concatenate them to obtain the composition visual proxies, $v^c$:
\begin{equation}
v_{i,j}^c=E_c([v_i^a,v_j^o]),
\end{equation}
where $E_c$ is a fully connected layer that projects the concatenated composition proxies into the visual embedding space, ensuring consistency with the dimensionality of the image features.

% 为了在每个分支上建立视觉原型与图像特征的对齐，我们设计了一种基于MLP的图像特征分解方法。具体而言，对于构成分支，我们直接使用全局图像特征 $f_v^{cls}$ 作为构成图像特征 $f^c$。对于属性和对象分支，通过两个MLP模块（\textbf{AD-MLP} 和 \textbf{OD-MLP}）将构成图像特征 $f^c$ 转化得到分解的图像特征。
\noindent
\textbf{MLP Decoupling Module.} To decompose features in the visual modality for visual proxies learning, we replace the Cross-Modal Decoupling Module with an MLP Decoupling Module because it is an intra-modality learning process.
% Just like textual prototypes, visual prototypes also need to align with decomposed image features. However, what differentiates visual prototypes from textual prototypes is that visual prototypes and image features both belong to the visual modality, and image features do not need to be projected onto the textual modality. Therefore, for visual prototypes, we designed an image feature decomposition method based on MLP. 
As described in \cref{Basic Framework}, for the attribute and object branches, the global image feature \( f_v^{cls} \) is processed through the \textbf{AD-MLP} and \textbf{OD-MLP} modules to obtain the attribute image feature \( f_v^a \) and object image feature \( f_v^o \), while the composition image feature remains unchanged, i.e., \( f_v^c = f_v^{cls} \).

% Specifically, for the composition branch, we directly use the global image feature $f_v^{cls}$ as the composition image feature $f^c$. For the attribute and object branches, the decomposed image features are derived by transforming the composition image feature $f^c$ through two MLP modules, \textbf{AD-MLP} for attributes and \textbf{OD-MLP} for objects, denoted as:
% \begin{equation}
%     f_v^a=MLP(f_c), \quad  f_v^o=MLP(f_c).
% \end{equation}

% 原型训练。我们通过计算图像特征与视觉原型之间的余弦相似度来聚类同一类别的图像特征以形成视觉原型，基于视觉原型，我们如下推导属性、对象和构成分支的概率：
\noindent
\textbf{Visual Proxies Training.} 
To construct discriminative visual proxies, we optimize feature embeddings via intra-category attraction and inter-category repulsion, the specific probability calculations for each branch are outlined as follows: 
\begin{equation}
p_v(y_{i,j}|x) = \frac{\exp(f_v^y \cdot v_{i,j}^y / \tau_v)}{\sum_{k=1}^{|\mathcal{Y}|} \exp(f_v^y \cdot v_k^y / \tau_v)},
\end{equation}
where $y \in \{{a,o,c}\}$, $|\mathcal{Y}| \in \{{|\mathcal{A}|,\mathcal|{O}|,|\mathcal{C}_s|}\}$, \( y_{i,j} \) represents the specific element in the set for each case (i.e., \( a_i \), \( o_j \), or \( c_{i,j} \)), $\tau_v$ is the temperature coefficient for the visual modality.

% \begin{equation}
% p_v({a_i}|x)=\frac{\exp(f^c \cdot v_i^a/\tau)}{\sum_{k=1}^{|\mathcal{A}|}\exp({f^c}\cdot v_k^a/\tau)},
% \end{equation}
% \begin{equation}
% p_v({o_j}|x)=\frac{\exp(f^c \cdot v_j^o/\tau)}{\sum_{k=1}^{|\mathcal{O}|}\exp({f^c}\cdot v_k^o/\tau)},
% \end{equation}
% \begin{equation}
% p_v(c_{i,j}|x) = \frac{\exp(f^c \cdot v_{i,j}^c/\tau)}
% {\sum_{k=1}^{|\mathcal{Y}_s|}\exp(f^c \cdot v_k^c/\tau)}
% \end{equation}
Subsequently, we calculate the cross-entropy loss for each branch:
\begin{equation}
\mathcal{L}_v^y = -\frac{1}{|\mathcal{X}|} \sum_{x \in \mathcal{X}} \log p_v(y | x),
\end{equation}
where $y \in \{{a,o,c}\}$. This contrastive objective enforces image features to cluster tightly around their assigned proxies (intra-category compactness) while pushing away irrelevant proxies (inter-category separability).

% \begin{equation}
%     \mathcal{L}_v^a =-\frac1{|\mathcal{X}|}\sum_{x\in\mathcal{X}}\log p_v(a|x), 
% \end{equation}
% \begin{equation}
%     \mathcal{L}_v^o =-\frac1{|\mathcal{X}|}\sum_{x\in\mathcal{X}}\log p_v(o|x),
% \end{equation}
% \begin{equation}
%     \mathcal{L}_v^c =-\frac1{|\mathcal{X}|}\sum_{x\in\mathcal{X}}\log p_v(c|x),
% \end{equation}
Therefore, the total loss for the visual proxies learning module $\mathcal{L}_v$ is defined as:
\begin{equation}
\mathcal{L}_v=\gamma_{ao}(\mathcal{L}_v^a+\mathcal{L}_v^o)+\gamma_{c}\mathcal{L}_v^c,
\end{equation}
the analysis of the hyperparameters \( \gamma_{ao} \) and \( \gamma_{c} \) can be found in the supplementary material.
\subsection{Cross-Modal Joint Learning}
% \textbf{Training. }为同时学到最优文本原型和视觉原型，我们引入额外的约束来对齐双模态原型。考虑到文本原型较为成熟，但视觉原型是从头开始训练的，容易学偏。我们采用KL散度，一种非对称的分布度量方式，对两个原型的分布进行约束。其中文本原型作为目标分布，视觉原型为近似分布，在优化过程中，通过最小化
\textbf{Training.} To jointly learn optimal textual prototypes and visual proxies, we propose a KL divergence-based distribution alignment framework. Since the textual prototypes inherently capture semantically stable class representations through linguistic priors, while visual proxies trained from scratch are susceptible to training biases, we designate the textual prototypes as the target distribution and the visual proxies as the approximate distribution. The loss function is:
% To obtain optimal textual and visual prototypes simultaneously, the textual prototypes and visual prototypes are aligned with the transformed image features. Additionally, these two prototypes need to be aligned with each other to ensure their concurrent effectiveness during the testing phase. Considering that the text prototypes is more mature while the visual prototypes is trained from scratch and prone to learning bias, we use the Kullback-Leibler (KL) divergence to constrain the distributions of the two prototypes, bring the visual prototypes distribution closer to the text prototypes distribution to learn the optimal visual prototypes.
\begin{equation}
    \mathcal{L}_{kl}=D_{KL}(P_t\|P_v)=\sum_xP_t(x)\log\frac{P_t(x)}{P_v(x)}
\end{equation}
where $P_t$ represents the probability distribution estimated by the textual prototypes, while $P_v$ represents the probability distribution estimated by the visual proxies. Consequently, the total loss $\mathcal{L}$ is expressed as follows:
\begin{equation}
    \mathcal{L}=\alpha(\mathcal{L}_t+\mathcal{L}_v)+\beta\mathcal{L}_{kl},
\end{equation}
the analysis of the hyperparameters \( \alpha \) and \( \beta \) can be found in the supplementary material.

By minimizing the loss function, we can obtain the optimal textual prototypes and visual proxies:
\begin{equation}
\mathbf{t}^*, \mathbf{v}^* = \arg\min_{\mathbf{t}, \mathbf{v}} \left( \mathcal{L} \right)
\end{equation}

\noindent
\textbf{Inference.} The trained textual and visual proxies are fused through multimodal probability aggregation. For each branch (attribute, object and composition), the final prediction probability is computed by:
\begin{equation}
p(y_{i,j}|x) = p_t(y_{i,j}|x) + \lambda p_v(y_{i,j}|x), 
\end{equation}
% \begin{equation}
% p({a_i}|x)=\alpha p_t({a_i}|x)+(1-\alpha)p_v({a_i}|x),
% \end{equation}
% \begin{equation}
% p({o_j}|x)=\alpha p_t({o_j}|x)+(1-\alpha)p_v({o_j}|x),
% \end{equation}
% \begin{equation}
% p({c_{i,j}}|x)=\alpha p_t({c_{i,j}}|x)+(1-\alpha)p_v({c_{i,j}}|x),
% \end{equation}
where $y \in \{a, o, c\}$, $\lambda$ balances cross-modal contributions (empirically set to 1 for simplicity). The probability of the final prediction for a given sample is calculated as the sum of the attribute, object and composition probability:
\begin{equation}
{p'}(c_{i,j}|x)=p(c_{i,j}|x)+p(a_i|x)+p(o_j|x).
\end{equation}
The final predicted composition is:
\begin{equation}
    y'=\arg\max\limits_{c_{i,j}\in\mathcal{C}_{pred}}(p'(c_{i,j}|x))
\end{equation}
\section{Experiments}
\subsection{Experiments Setting}

\textbf{Datasets.} We evaluated the model's performance on four datasets: UT-Zappos \cite{ut}, MIT-States \cite{mit}, C-GQA \cite{cgqa}, and VAW-CZSL\cite{saini2022disentangling}. The dataset information is shown in the supplementary material.

% UT-Zappos is a large shoe dataset consisting of 16 attributes and 12 objects. MIT-States is a diverse collection of everyday objects, featuring 115 attributes and 245 objects. C-GQA is the largest dataset for the CZSL task, derived from the GQA dataset \cite{gqa}, containing 453 attributes and 870 objects. We followed the dataset split standards from previous studies\cite{csp23,naeem2021learning} and the statistics are provided in \cref{dataset}.

% \begin{table}[htbp]
% \setlength\tabcolsep{2.2pt}
% \centering
% \caption{The statistics of three CZSL datasets.}
% %\vspace{-0.3cm}
% \small
% \begin{tabular}{c|cc|cc|ccc|ccc}
% \hline
% \hline
% \multirow{2}{*}{Dataset} & \multicolumn{2}{c|}{} & \multicolumn{2}{c|}{\textbf{Train}} & \multicolumn{3}{c|}{\textbf{Val}} & \multicolumn{3}{c}{\textbf{Test}} \\
% \cline{2-11}
%                         & $\mathcal{A}$ & $\mathcal{O}$ & $\mathcal{Y}_s$ & $\mathcal{X}$ & $\mathcal{Y}_s$ & $\mathcal{Y}_u$ & $\mathcal{X}$ & $\mathcal{Y}_s$ & $\mathcal{Y}_u$ & $\mathcal{X}$ \\
% \hline
% MIT-States           &115 & 245 & 1.2k & 30k & 300 & 300 & 10k & 400 & 400 & 13k \\
% \hline
% UT-Zappos              &16 & 12 & 83 & 23k & 15 & 15 & 3k & 18 & 18 & 3k \\
% \hline
% C-GQA          &413 & 674 & 5.5k & 27k & 1.2k & 1k & 7k & 888 & 923 & 5k \\
% \hline
% \hline
% \end{tabular}
% \label{dataset}
% \end{table}

\noindent
\textbf{Metrics.} Following standard open/closed-world settings \cite{li2024context,trokia24}, we evaluate using Seen (\textbf{S})/Unseen (\textbf{U}) composition accuracy, Harmonic Mean (\textbf{HM}), and Area Under Curve (\textbf{AUC}).
\begin{table*}[t!]
\centering
\caption{The experimental results for both closed/open-world settings. The best performance are highlighted in bold.}
% \footnotesize
%\vspace{-0.3cm}
\resizebox{0.8\linewidth}{!}{
\begin{tabular}{r|c|cccc|cccc|cccc}
\hline
\hline
\multirow{2}{*}{\textbf Method} & \multirow{2}{*}{\textbf Venue} & \multicolumn{4}{c|}{C-GQA} & \multicolumn{4}{c|}{UT-Zappos} & \multicolumn{4}{c}{MIT-States} \\ \cline{3-14}
                       &  & S    & U    & HM   & AUC    &   S    & U    & HM   & AUC    &   S    & U    & HM   & AUC  \\ \hline
\multicolumn{14}{c}{Closed-world Results}    \\                                                                               \hline
CLIP\cite{clip}  &  ICML'21     & 7.5  & 25.0 & 8.6  & 1.4   & 15.8 & 49.1 & 15.6 & 5.0   & 30.2 & 46.0 & 26.1 & 11.0  \\
CoOp\cite{coop}  &  IJCV'22     & 20.5 & 26.8 & 17.1 & 4.4   & 52.1 & 49.3 & 34.6 & 18.8  & 34.4 & 47.6 & 29.8 & 13.5  \\
CSP\cite{csp23}  &  ICLR'23       & 28.8 & 26.8 & 20.5 & 6.2   & 64.2 & 66.2 & 46.6 & 33.0  & 46.6 & 49.9 & 36.3 & 19.4  \\
% PCVL\cite{PCVL}        &  arXiv'22      & 48.5 & 47.2 & 35.3 & 18.3   & 64.4 & 64.0 & 46.1 & 32.2  &   -  & -    & -    & -    \\
DFSP(i2t)\cite{DFSP}  &  CVPR'23         & 35.6 & 29.3 & 24.3 & 8.7   & 64.2 & 66.4 & 45.1 & 32.1  & 47.4 & 52.4 & 37.2 & 20.7  \\
DFSP(BiF)\cite{DFSP}  &  CVPR'23         & 36.5 & 32.0 & 26.2 & 9.9   & 63.3 & 69.2 & 47.1 & 33.5  & 47.1 & 52.8 & 37.7 & 20.8  \\
DFSP(t2i)\cite{DFSP}  &  CVPR'23         & 38.2 & 32.0 & 27.1 & 10.5  & 66.7 & 71.7 & 47.2 & 36.0  & 46.9 & 52.0 & 37.3 & 20.6  \\
DLM\cite{hu2024dynamic}  &  AAAI'24               & 32.4 & 28.5 & 21.9 & 7.3   & 67.1 & 72.5 & 52.0 & 39.6  & 46.3 & 49.8 & 37.4 & 20.0  \\ 
ProLT\cite{ProLT}  &  AAAI'24             & 39.5 & 32.9 & 27.7 & 11.0  & 66.0 & 70.1 & 49.4 & 36.1  & 49.1 & 51.0 & 38.2 & 21.1  \\ 
PLID\cite{PLID}   & ECCV'24   & 38.8 & 33.0 & 27.9 & 11.0  & 67.3 & 68.8 & 52.4 & 38.7  & 49.7 & 52.4 & 39.0 & 22.1  \\
CDS-CZSL\cite{li2024context}  &  CVPR'24          & 38.3 & 34.2 & 28.1 & 11.1  & 63.9 & 74.8 & 52.7 & 39.5  & 50.3 & 52.9 & 39.2 & 22.4  \\ 
Troika\cite{trokia24}  &  CVPR'24            & 41.0 & 35.7 & 29.4 & 12.4  & 66.8 & 73.8 & 54.6 & 41.7  & 49.0 & \textbf{53.0} & 39.3 & 22.1  \\ 
% Retrieval-Augmented
% & 50.0 & 53.3 & 39.2 & 22.5   &   69.4 & 72.8 & 56.5 & 44.5   &   45.6 & 36.0 & 32.0 & 14.4 &     \\
IMAX\cite{10737702}                   & TPAMI'25 &39.7 & 35.8& 29.8 & 12.8 & 69.3&70.7 & 54.2 & 40.6 & 48.7& 53.8  & 39.1 & 21.9      \\ 
\textbf{VP-CMJL(Ours)}   &  &   \textbf{46.0} & \textbf{40.2} & \textbf{34.9} & \textbf{16.3}   &   \textbf{71.9} & \textbf{76.3} & \textbf{58.5} & \textbf{47.9}  &   \textbf{51.8} & 52.6 & \textbf{40.4} & \textbf{23.3} \\ \hline
\multicolumn{14}{c}{Open-world Results}                                                                                    \\ \hline
CLIP\cite{clip} &    ICML'21        & 7.5  & 4.6  & 4.0  & 0.3    &   15.7 & 20.6 & 11.2 & 2.2    & 30.1 & 14.3 & 12.8 & 3.0 \\
CoOp\cite{coop} &    IJCV'22        & 21.0 & 4.6  & 5.5  & 0.7    &   52.1 & 31.5 & 28.9 & 13.2   & 34.6 & 9.3  & 12.3 & 2.8 \\
CSP\cite{csp23}  &    ICLR'23        & 28.7 & 5.2  & 6.9  & 1.2    &   64.1 & 44.1 & 38.9 & 22.7   & 46.3 & 15.7 & 17.4 & 5.7 \\
% PCVL\cite{PCVL} &    arXiv'22       & -    & -    & -    & -      &   64.6 & 44.0 & 37.1 & 21.6   & 48.5 & 16.0 & 17.7 & 6.1 \\
DFSP(i2t)\cite{DFSP} &  CVPR'23     & 35.6 & 5.6  & 9.0  & 1.9    &   64.3 & 53.8 & 41.2 & 26.4   & 47.2 & 18.2 & 19.1 & 6.7 \\
DFSP(BiF)\cite{DFSP}  & CVPR'23     & 36.5 & 7.6  & 10.6 & 2.4    &   63.5 & 57.2 & 42.7 & 27.6   & 47.1 & 18.1 & 19.2 & 6.7 \\
DFSP(t2i)\cite{DFSP}  &  CVPR'23    & 38.2 & 7.2  & 10.4 & 2.4    &   66.8 & 60.0 & 44.0 & 30.3   & 47.5 & 18.5 & 19.3 & 6.8 \\
PLID\cite{PLID}   & ECCV'24   & 39.1 & 7.5 & 10.6 & 2.5 & 67.6 & 55.5 & 46.6 & 30.8 & 49.1 & 18.7 & 20.4 & 7.3\\
CDS-CZSL\cite{li2024context} & CVPR'24      & 37.6 & 8.2 & 11.6 & 2.7 & 64.7 & 61.3 & 48.2 & 32.3 & 49.4 & \textbf{21.8} & \textbf{22.1} & \textbf{8.5} \\
Troika\cite{trokia24}     &  CVPR'24   & 40.8 & 7.9 & 10.9 & 2.7   &   66.4 & 61.2 & 47.8 & 33.0   & 48.8 & 18.4 & 20.1 & 7.2 \\
IMAX\cite{10737702}                  & TPAMI'25 &38.7 & 7.9& 11.2 & 2.5 & 68.4&57.3 & 47.5 & 32.3 & 50.2& 18.6  & 21.4 & 7.6      \\ 
\textbf{VP-CMJL(Ours)}   &  &   \textbf{46.0} & \textbf{11.5} & \textbf{15.5} & \textbf{4.6}& \textbf{71.9} & \textbf{66.6} & \textbf{54.5} & \textbf{41.4} & \textbf{51.8} & 19.9 & 22.0 & 8.3\\
\hline
\hline
\end{tabular}}
%}
\label{result}
\end{table*}

\begin{table}[t!]
\label{vaw}
\centering
\caption{Closed-world experimental results on the VAW-CZSL dataset, with the best performance in bold.}
% \footnotesize
\resizebox{0.9\linewidth}{!}{
%\vspace{-0.3cm}
\begin{tabular}{r|c|cccc}
\hline
\hline
\multirow{2}{*}{\textbf{Method}} & \multirow{2}{*}{\textbf{Venue}} & \multicolumn{4}{c}{VAW-CZSL} \\ \cline{3-6}
                       &  & S    & U    & HM   & AUC    \\ \hline
CLIP\cite{clip}  &  ICML'21     & 23.9  & 18.0 & 11.9  & 2.6   \\
CSP\cite{csp23}  &  ICLR'23     & 31.9 & 33.6 & 23.3 & 8.5   \\
DFSP\cite{DFSP}  &  CVPR'23  & 40.1 & 40.9 & 31.1 & 14.1   \\
CAILA\cite{DFSP}  &  WACV'24  & 41.6 & 49.2 & 34.6 & 17.2   \\
\textbf{VP-CMJL(Ours)}   &  & \textbf{47.8} & \textbf{51.1} & \textbf{38.2} & \textbf{20.7} \\ \hline
\hline
\end{tabular}}
\label{vaw}
\end{table}

% \textbf{Implementation Details.}为了公平起见，我们跟随前人的设定【】，选用图文预训练模型CLIP ViT-L【】作为编码器，并在图像编码器端引入参数微调模块adapter，针对于交叉注意力解耦模块，注意力头数设置为16，层数为1，维度为768。在训练阶段，全部采用adam优化器和StepLR学习率调度器，每三个epoch学习率衰减为原来的0.5。其中，针对于ut数据集，batch size设置为64，选用5×10的负四次方学习率，1×10的负五次方weight decay，超参数alpha设置为0.7，beta设置为0.3；针对于mit这个数据集，batch_size设置为32，选用5×10的负四次方学习率，1×10的负五次方weight decay，超参数alpha设置为0.3，beta设置为0.7；针对于cgqa数据集，batch size设置为16，选用选用5×10的负5次方学习率，1×10的负五次方weight decay，超参数alpha设置为0.5，beta设置为0.5。总共训练20个epoch，整体训练和测试都在NVIDIA A800 GPU上完成。
% cgqa选用5×10负5学习率。
\noindent
% \textbf{Implementation Details.} To ensure fairness, we adopt the parameter settings established by previous research, utilizing the pre-trained CLIP ViT-L model \cite{clip} as our image/text encoder. During training, we use the Adam optimizer in conjunction with a StepLR learning rate scheduler, where the learning rate decays by a factor of 0.5 every 3 epochs. For the UT-Zappos and MIT-States datasets, the learning rate is \(5 \times 10^{-4}\) and weight decay is \(1 \times 10^{-5}\); for the C-GQA dataset, the learning rate is \(5 \times 10^{-5}\) with the same weight decay. Training is conducted for 20 epochs in total. All training and testing are conducted on NVIDIA A800 GPUs.
\textbf{Implementation Details.} To ensure fairness, we adopt the parameter settings established by previous research, utilizing the pre-trained CLIP ViT-L/14 model \cite{clip} as our image/text encoder. Training is conducted for 20 epochs in total. All training and testing are conducted on NVIDIA A800 GPUs. More details can be found in the supplementary material.

\subsection{Main Results}
We comprehensively evaluate our model under both closed-world and open-world settings, comparing it with recent CLIP-based CZSL methods from the past two years, including: CLIP\cite{clip}, CoOp\cite{coop}, CSP\cite{csp23}, DFSP\cite{DFSP}, DLM\cite{hu2024dynamic}, ProLT\cite{ProLT}, PLID\cite{PLID}, CDS-CZSL\cite{li2024context}, Trokia\cite{trokia24} and IMAX\cite{10737702}. The results are presented in \cref{result}. In the closed-world setting, our \textbf{VP-CMJL} achieves SOTA performance across all three datasets on nearly all metrics. Specifically, \textbf{VP-CMJL} improves the HM by \( +5.5\% \), \( +3.9\% \), and \( +1.1\% \), and the AUC by \( +3.9\% \), \( +6.2\% \), and \( +1.2\% \), respectively, across the three datasets, demonstrating the effectiveness of the proposed method. In the open-world setting, \textbf{VP-CMJL} significantly outperforms competing methods on the UT-Zappos and C-GQA datasets, with HM improvements of \( +6.7\% \) and \( +4.6\% \), and AUC improvements of \( +8.4\% \) and \( +1.9\% \), respectively.
For the MIT-States dataset containing substantial label noise \cite{atzmon2020causal}, the baseline method CDS-CZSL adopts dedicated pruning techniques specifically designed for open-world settings. Our method achieves competitive performance without relying on these task-specific optimizations.

Beyond standard benchmarks, we conduct a comprehensive evaluation on VAW-CZSL, a new large-scale real-world attribute dataset. While most existing methods lack validation on this challenging benchmarks. As shown in \cref{vaw}, our approach improves the HM by +3.6\% and AUC +3.5\%, demonstrating the generalization ability and effectiveness of our model.

\begin{figure*}[htbp]
  \centering
  \includegraphics[width=0.9\textwidth]{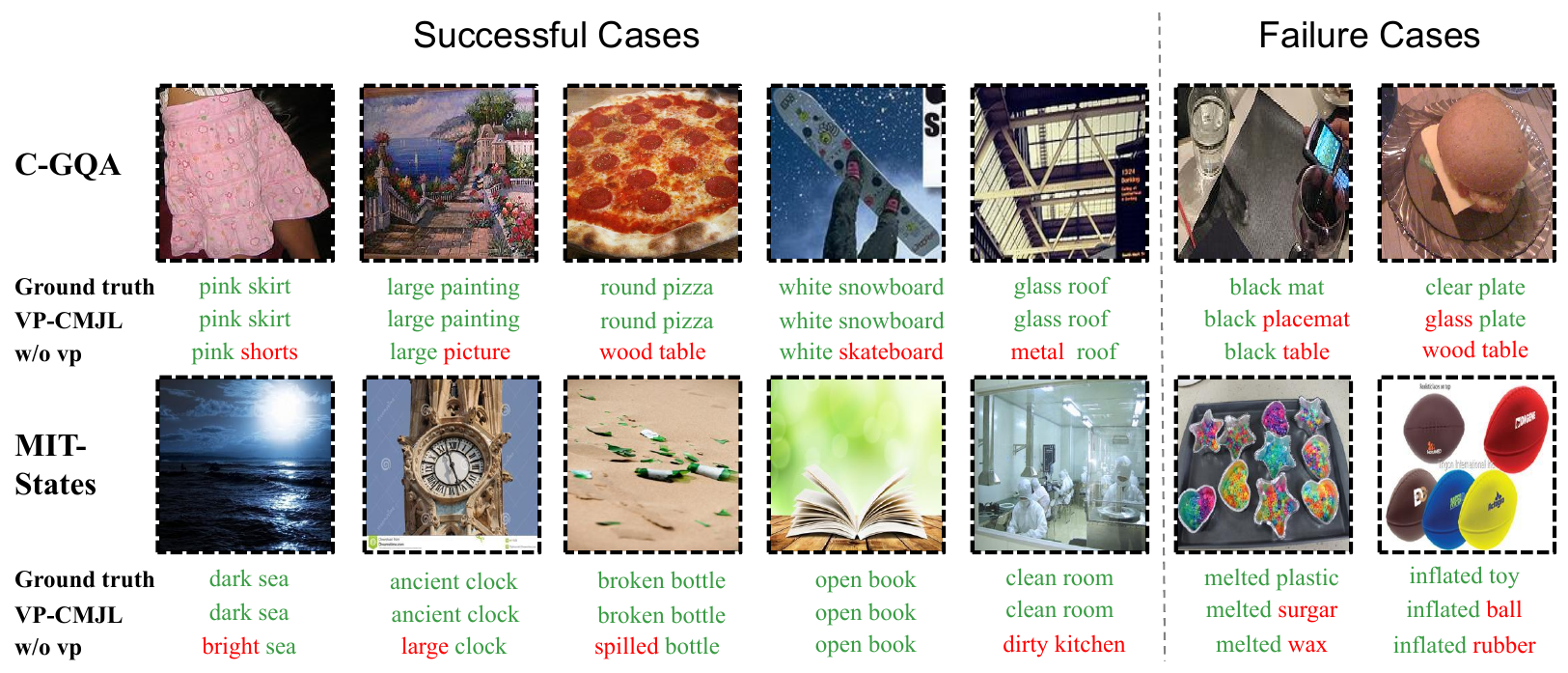}
  \caption{Qualitative results. The term 'w/o VP' refers to the text-prototype-based method, while the green font indicates correct labels and the red font indicates incorrect labels.}
  
  \label{sample}
  
  %\vspace{-0.5cm}
\end{figure*}

\subsection{Ablation Study}
\textbf{Ablation study on \textbf{VP-CMJL}.} To validate the effectiveness of visual proxies and cross-modal joint learning, we conduct ablation experiments in a closed-world setting on the UT-Zappos and MIT-States datasets. The experimental setup is as follows: (1) \textbf{Component removal}: Remove either the textual prototype (TP) or visual proxy (VP) during training and inference; (2) \textbf{Single-modal removal during inference}: Remove TP/VP only during testing, while maintaining full dual-modal training. The results (\cref{ablation1}) show: (1) \textbf{Component necessity}: Removing either component causes a significant performance drop, confirming that the introduction of visual proxies benefits the CZSL task and must be learned in conjunction with textual prototypes, enhancing both CLIP's generalization and fine-grained discrimination capabilities. (2) \textbf{Training-testing dynamic difference}: The performance degradation is smaller when removing a modality during testing than during training, indicating that joint optimization promotes a synergistic representation enhancement mechanism during training.
\begin{table}[htbp]
\centering
\caption{Ablation experiment results on dual-modal center learning. 'TP'/'VP' refers to text prototypes and visual proxies.}
%\vspace{-0.3cm}
% \footnotesize
\resizebox{0.9\linewidth}{!}{
\begin{tabular}{cc|cccc|cccc}
\hline
\hline
\multicolumn{2}{c|}{ } & \multicolumn{4}{c|}{UT-Zappos} & \multicolumn{4}{c}{MIT-States} \\ \cline{1-10} 
 TP & VP & S & U & HM & AUC & S & U & HM & AUC \\ \hline
 \multicolumn{10}{c}{Training and inference}                                              \\ \hline
\checkmark & \checkmark & \textbf{71.9} & \textbf{76.3} & \textbf{58.5} & \textbf{47.9} & \textbf{51.8} & \textbf{52.6} & \textbf{40.4} & \textbf{23.3} \\ 
\checkmark &  \ding{55} & 64.4 & 70.7 & 51.9 & 37.8 & 47.1 & 52.1 & 37.8 & 20.8 \\ 
\ding{55} & \checkmark & 65.8 & 72.3 & 55.3 & 42.1 & 50.5 & 49.4 & 37.6 & 20.7 \\ 
% \ding{55} & \ding{55} & 65.2 & 42.4 & 42.4 & 24.2 & 51.1 & 38.4 & 33.2 & 16.5  \\ 
\hline
 \multicolumn{10}{c}{Inference}                                              \\ \hline
\checkmark & \checkmark & \textbf{71.9} & \textbf{76.3} & \textbf{58.5} & \textbf{47.9} & \textbf{51.8} & \textbf{52.6} & \textbf{40.4} & \textbf{23.3} \\ 
\checkmark &  \ding{55} & 69.2 & 75.1 & 56.7& 44.2 & 51.3 & 52.3 & 40.4 & 23.0 \\ 
\ding{55} & \checkmark & 69.5 & 76.4 & 57.8 & 46.4 & 48.7 & 50.6 & 37.3 & 20.4  \\ 
\hline
\hline
\end{tabular}}
\label{ablation1}
\end{table}

\noindent
\textbf{Ablation study on decoupling modules.} We exchange the decoupling modules on both modality. As shown in \cref{ablation2}, the CA-/MLP-based decoupling module achieves the best effect for textual prototpyes and visual proxies learning respectively. The results confirm that cross-modal decomposition aids in modality alignment, while MLP-based transformation is more effective for intra-modal learning.

\setlength\tabcolsep{3pt}
\begin{table}[htbp]
%\vspace{-0.15cm}
\small
\centering
\caption{Results of ablation experiments on the decomposition modules. i2t/i2v represents the image feature decoupling method used in textual prototype/visual proxy learning, respectively.}
%\vspace{-0.3cm}
% \footnotesize
\resizebox{0.9\linewidth}{!}{
\begin{tabular}{cc|cccc|cccc}
\hline
\hline
\multicolumn{2}{c|}{ } & \multicolumn{4}{c|}{UT-Zappos} & \multicolumn{4}{c}{MIT-States} \\ \cline{1-10} 
i2t & i2v & S & U & HM & AUC & S & U & HM & AUC \\ \hline
CA & MLP & \textbf{71.9} & \textbf{76.3} & \textbf{58.5} & \textbf{47.9} & \textbf{51.8} & \textbf{52.6} & \textbf{40.4} & \textbf{23.3} \\ 
CA & CA & 67.1 & 75.2 & 54.7 & 42.0 & 50.9 & 51.6 & 39.6 & 22.3 \\ 
MLP& CA & 67.3 & 76.2 & 55.7 & 44.2 & 50.8 & 52.5 & 39.6 & 22.8 \\ 
MLP & MLP & 69.5 & 73.1 & 58.5 & 45.3 & 50.0 & 51.4 & 38.8 & 21.8 \\ 
\hline
\hline
\end{tabular}}
\label{ablation2}
\end{table}

\subsection{Quality Analysis}
% 我们在图三中可视化模型在CGQA和MIT数据集上的定性结果。具体来说，我们展示了所提出的模型DMLP和仅基于文本原型的多路预测方法的部分成功预测案例和失败案例。由结果可见，DMLP可以准确判别相似组合对，比如large painting和large picture，glass roof和metal roof等，而仅基于文本原型的方法并不能很好的图像外观较为相似的组合对。这表明，我们的模型学习到了类别的细粒度特征。
\textbf{Case analysis.}
We visualize the qualitative results of the model on the C-GQA and MIT-States datasets in \cref{sample}, showcasing both successful and failure cases for the proposed \textbf{VP-CMJL} model, as well as the text-prototype-based method ('w/o VP'). The results clearly show that \textbf{VP-CMJL} effectively distinguishes between similar compositions, such as "large painting" vs. "large picture" or "broken bottle" vs. "spilled bottle," while the text-prototype-based method struggles with compositions that have similar visual appearances. This demonstrates \textbf{VP-CMJL}'s ability to learn fine-grained composition features. In failure cases, although the model often misclassifies the composition, it typically correctly identifies one of the primitives. Misclassifications are generally due to semantic similarity to the true label or some ambiguity.

\begin{figure}[htbp]
  \centering
   \includegraphics[width=0.9\linewidth]{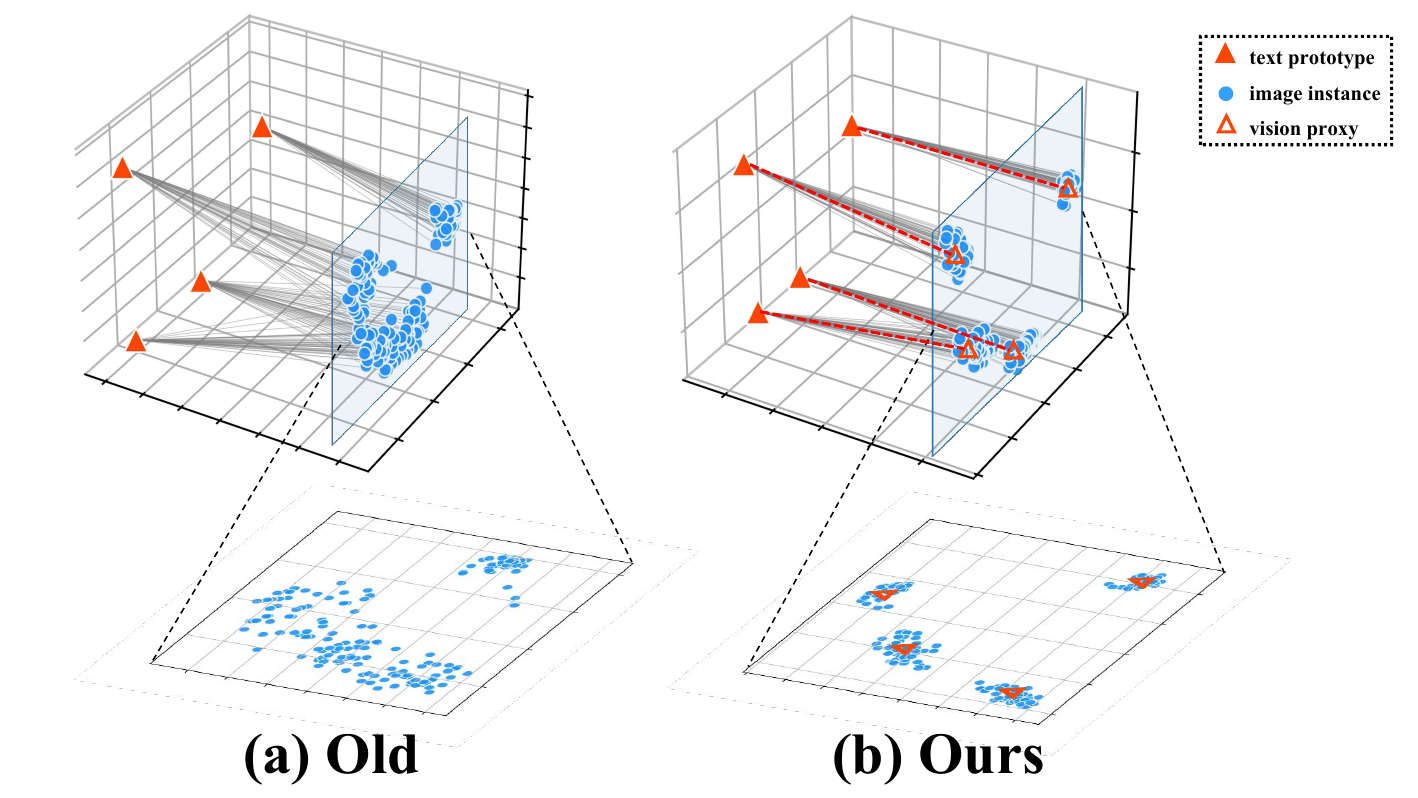}
   \caption{(a) Example of modality space feature distribution in traditional three-path methods. (b) Example of modality space feature distribution in \textbf{VP-CMJL}.}
   \label{fig:3d}
   %\vspace{-0.5cm}
\end{figure}

\textbf{Visualization of modality space features.}
As shown in \cref{fig:3d}, the comparison of modality feature spaces between the traditional three-path method and the proposed \textbf{VP-CMJL} framework highlights key differences. The baseline method suffers from rigid matching between textual prototypes and image features, resulting in cross-modal similarity measurement bias and a dispersed visual feature space with overlapping class boundaries. In contrast, \textbf{VP-CMJL} introduces a learnable visual proxy and bidirectional cross-modal adjustment, achieving semantic alignment and dynamically adjusting the mapping from image features to textual prototypes, leading to a more compact visual feature space and better class separability.
% \setlength\tabcolsep{3pt}
% \begin{table}[htbp]
% \centering
% \small
% \begin{tabular}{c|cccc|cccc}
% \hline
% \multirow{2}{*}{Model } & \multicolumn{4}{c|}{UT-Zappos} & \multicolumn{4}{c}{MIT-States} \\ 
%      & S & U & HM & AUC & S & U & HM & AUC \\ \hline
%  CLIP& \textbf{69.6} & \textbf{77.53} & \textbf{57.99} & \textbf{46.83} & \textbf{51.8} & \textbf{52.6} & \textbf{40.4} & \textbf{23.27} \\ 
%  Bert&  &  &  &  &  &  &  &  \\ 
%  GPT&  & &  &  &  &  &  &  \\ 
% \hline
% \end{tabular}
% \caption{Results on UT-Zappos and MIT-States datasets Visual prototype with different initializations.}
% \label{table:results}
% \end{table}

\section{Conclusion}
% 在本文中，我们考虑到视觉模态对组合零样本学习具备天然的优势，首次在视觉模态引入视觉原型，并结合文本原型，提出一种双模态原型联合学习方法。该方法采用属性、对象和组合三路预测框架，针对每个模态原型设计特定的图像特征解耦方法，利用联合学习策略，以获得具备广泛概念的文本原型和具备细粒度特征的视觉原型。最终，利用这两个模态原型作为最终分类的依据，从而提高czsl任务在不可见对的泛化能力和对相似组合对的判别能力。通过模型评估消融实验，我们充分证明了双模态原型学习的有效性，并在三个数据集上实现了sota性能。我们希望我们的工作为CZSL任务打开双模态原型学习的新思路。
% In this paper, we propose a novel method for the CZSL task. Recognizing the inherent advantages of the visual modality in CZSL, we introduce visual prototypes to enhance the model’s ability to capture fine-grained information in the visual space. We also develop tailored decomposition modules and a joint learning strategy to enhance feature representation, allowing the model to optimize prototypes across both modalities for the first time. These prototypes capture essential category information during training and act as crucial reference points during inference. Our experimental results demonstrate state-of-the-art performance in the closed-world setting and competitive results in the open-world setting across three public CZSL datasets, showcasing the effectiveness of our proposed method.
This work introduces the Visual Proxy and Cross-Modal Joint Learning strategy to address the challenges of Compositional Zero-Shot Learning (CZSL). By tackling the modality gap and the lack of fine-grained cues, our method introduces visual proxies for the first time and proposes a cross-modal joint learning strategy that enhances modality alignment and improves the discrimination of similar compositions, boosting generalization in CZSL tasks. Extensive experiments demonstrate that \textbf{VP-CMJL} outperforms existing methods in both closed-world and open-world settings, achieving state-of-the-art results. This work presents a promising direction for enhancing the generalization and discriminative power of compositional zero-shot learning by combining the strengths of textual and visual modalities.

\section*{Acknowledgements}
This work is supported by the National Key R\&D Program of China (No. 2023YFC3805203), National Natural Science Foundation of China (No. 62402354), and the Haikou Science and Technology Plan Project (2023-053).

{
    \small
    \bibliographystyle{ieeenat_fullname}
    \bibliography{main}
}
\end{document}